%% file: Template.tex
\documentclass{article}
\usepackage{spconf,amsmath,graphicx,hyperref}
\usepackage{times}
\usepackage{helvet}
\usepackage{caption}
\usepackage{algorithm}
\usepackage{algorithmic}
\usepackage{amsfonts}
\usepackage{bm}
\usepackage{mathrsfs}
\usepackage{xcolor,colortbl}

\usepackage{subfigure}
\usepackage{arydshln,booktabs,multirow}
\usepackage{bigstrut,bigdelim}
\usepackage{paralist}

\usepackage{makecell}
\usepackage{nicefrac}
\usepackage{epsfig}
\usepackage{diagbox}
\usepackage{framed}
\usepackage{empheq}
\usepackage{array}
\usepackage{enumitem}
\usepackage{mdwlist}
\usepackage{xspace,mfirstuc,tabulary}
\usepackage{tcolorbox}
\usepackage{algorithm}
\usepackage{amssymb}
\usepackage{mathtools}
\usepackage{amsthm}
\usepackage{subfigure}
\usepackage{placeins}

\title{FGGM: Fisher-Guided Gradient Masking for Continual Learning}
\name{\parbox{\linewidth}{\centering
    Chao-Hong Tan\qquad
    Qian Chen\qquad
    Wen Wang\qquad
    Yukun Ma\qquad
    Chong Zhang\qquad \\
    Chong Deng\qquad 
    Qinglin Zhang\qquad
    Xiangang Li\qquad
    Jieping Ye
}}
\address{Tongyi Lab, Alibaba Group \\ {\texttt{\{tanchaohong.ch, tanqing.cq, w.wang\}@alibaba-inc.com}}}

\begin{document}

\ninept
\maketitle
\begin{abstract}
  Catastrophic forgetting impairs the continuous learning of large language models. We propose \textbf{Fisher-Guided Gradient Masking (FGGM)}, a framework that mitigates this by strategically selecting parameters for updates using diagonal Fisher Information. FGGM dynamically generates binary masks with adaptive thresholds, preserving critical parameters to balance stability and plasticity without requiring historical data. Unlike magnitude-based methods such as MIGU, our approach offers a mathematically principled parameter importance estimation. On the TRACE benchmark, FGGM shows a 9.6\% relative improvement in retaining general capabilities over supervised fine-tuning (SFT) and a 4.4\% improvement over MIGU on TRACE tasks. Additional analysis on code generation tasks confirms FGGM's superior performance and reduced forgetting, establishing it as an effective solution.
\end{abstract}

\input{chapters/1-introduction}

\input{chapters/3-preliminary}
\input{chapters/4-method}

\input{chapters/5-experiments}

\section{Conclusion}
\vspace{-1mm}
We propose the FGGM framework for continual learning of LLMs, leveraging Fisher Information Matrix (FIM) based gradient masking to achieve parameter plasticity-stability balance with zero access to previous task data. 
Experimental results demonstrate that our method achieves state-of-the-art (SOTA) performance in maintaining stability among approaches requiring no historical data access. 
Next, we will enhance FGGM’s learning capabilities on downstream tasks, optimize its efficiency, and explore its applicability to multimodal fine-tuning scenarios.

\vfill\pagebreak

\bibliographystyle{IEEEbib}
\bibliography{custom_short}

\end{document}

%% file: chapters/1-introduction.tex
\section{Introduction}
  The challenge of catastrophic forgetting is exacerbated as large language models (LLMs) become widely adopted universal tools for diverse NLP tasks. Catastrophic forgetting refers to the tendency of neural networks to abruptly lose previously acquired knowledge when trained on new tasks or datasets. This phenomenon poses a significant challenge in the realm of continual learning (CL), where models are expected to learn and adapt to a stream of tasks over time without forgetting prior information. To address catastrophic forgetting, current mitigation strategies during continual learning broadly fall into three categories:
  (1) Replay-based methods, which retain~\cite{DBLP:conf/emnlp/ScialomCM22,DBLP:conf/naacl/JinZZ00WA022} or generate past data~\cite{DBLP:conf/iclr/SunHL20,DBLP:conf/iclr/QinJ22,DBLP:journals/pami/LiH18a} through experience rehearsal or generative models to mitigate catastrophic forgetting of old tasks. 
  (2) Parameters-freezing based methods, which focus on freezing base model parameters and only train a small set of additional parameters, including Prompt Tuning~\cite{DBLP:conf/emnlp/LesterAC21}, Prefix Tuning~\cite{DBLP:conf/acl/LiL20}, Low-Rank Adaptation (LoRA)~\cite{DBLP:conf/iclr/HuSWALWWC22,DBLP:conf/emnlp/WangCGXBZZGH23}, Adapters~\cite{DBLP:conf/icml/HoulsbyGJMLGAG19}, Mixture of Experts (MoE)~\cite{DBLP:conf/iclr/ShazeerMMDLHD17}, Model Expansion~\cite{DBLP:journals/corr/ChenGS15} such as LLaMA Pro~\cite{DBLP:conf/acl/WuGGLWFSL24}.
  (3) Regularization-based methods that penalize or mitigate changes to the weights that are important for previous trained tasks~\cite{DBLP:journals/corr/KirkpatrickPRVD16,DBLP:conf/emnlp/ChenHCCLY20,DBLP:conf/emnlp/DuCLQHCC024}, or methods that restrict directions of model updates such as orthogonal gradient 
  directions (OGD)~\cite{DBLP:conf/aistats/FarajtabarAML20}.

  However, replay-based methods necessitate storing historical data, posing challenges to privacy and scalability. Parameters-freezing approaches typically require freezing base models, potentially limiting downstream task performance. Among regularization-based methods, EWC~\cite{DBLP:journals/corr/KirkpatrickPRVD16} relies on access to historical data, while OGD~\cite{DBLP:conf/aistats/FarajtabarAML20} imposes excessive memory demands that hinder direct applications to full fine-tuning of LLMs. Recently, MIGU~\cite{DBLP:conf/emnlp/DuCLQHCC024} proposes a promising replay-free and task-label-agnostic method by selectively updating parameters with large output magnitudes in linear layers to mitigate catastrophic forgetting. MIGU outperforms full fine-tuning and competitive replay-based, parameters-freezing, and regularization-based methods on mitigating catastrophic forgetting.  However, MIGU relies on \textit{empirical observations} of output magnitudes and lacks theoretical foundation for parameter importance estimation. 
  
  Our work is motivated to address the limitations of MIGU. We propose a \textbf{Fisher-Guided Gradient Masking (FGGM) framework} that strategically selects parameters for updates. The Fisher Information Matrix has been demonstrated as an effective metric for estimating parameter importance by quantifying their contribution to task loss through gradient analysis~\cite{DBLP:journals/corr/KirkpatrickPRVD16}. FGGM 
  first estimates the Fisher information matrix using new task data, quantifies parameter significance, generates binary masks via dynamic thresholds, and dynamically controls gradient updates to alleviate catastrophic forgetting.
  In summary, our contributions are three-fold:
    \begin{itemize}[leftmargin=*,noitemsep]
    \item We propose a novel Fisher-Guided Gradient Masking (\textbf{FGGM}) framework without the need to access past data, for mitigating catastrophic forgetting in LLMs during continual learning.
    \item FGGM introduces a \textbf{theoretically grounded parameter importance estimation mechanism based on Fisher information}. It directly measures parameter-task relevance and enables dynamic gradient gating to balance \textbf{plasticity} (acquiring new capabilities) and \textbf{stability} (retaining pre-existing knowledge) of LLMs.
    \item Empirical validations on standard benchmarks demonstrate that FGGM outperforms MIGU by effectively addressing catastrophic forgetting without compromising model adaptability. These results underscore the effectiveness and versatility of principled gradient masking strategies for continual learning in LLMs.
    \end{itemize}

%% file: chapters/3-preliminary.tex
\section{Preliminaries}  
\label{sec:preliminaries}

\textbf{Continual learning (CL)}, also termed lifelong learning, addresses the challenge of training machine learning models on a non-stationary stream of tasks $\{ \mathcal{T}_1, \mathcal{T}_2, \dots, \mathcal{T}_N \}$. Each task $\mathcal{T}_t$ is associated with a data distribution $\mathcal{D}_t = \{x_i, y_i\}_{i=1}^{M_t}$, where $x_i$ denotes inputs, $y_i$ denotes task-specific targets, and $M_t$ is the dataset size for task $t$. The objective of continual learning is to optimize model parameters $\theta$ across all tasks while preserving performance on previous tasks. Formally, the CL objective minimizes:  
\begin{equation}  
    \mathcal{L}_{\text{CL}}(\theta) = \sum_{t=1}^N \mathbb{E}_{(x,y) \sim \mathcal{D}_t} [ \ell(f_\theta(x), y) ]  
\end{equation}
\noindent where $\ell(\cdot)$ is the task-specific loss function, and $f_\theta(x)$ denotes the model predictions. The key challenge arises from the \textit{catastrophic forgetting} phenomenon: when training on $\mathcal{T}_t$, the model disproportionately overwrites parameters critical for earlier tasks $\mathcal{T}_{1:t-1}$~\cite{wang2023trace}.

The \textbf{Fisher Information Matrix (FIM)}~\cite{DBLP:journals/corr/KirkpatrickPRVD16} for task $\mathcal{T}_t$ can be compactly expressed as follows:
\begin{equation}  
\begin{split}  
F^{(t)}(\theta) = \mathbb{E}_{\substack{(x,y)\sim \mathcal{D}_t}} \Biggl[ & \nabla_\theta \log p(y|x, \theta) 
\cdot \left( \nabla_\theta \log p(y|x, \theta) \right)^\top \Biggr] 
\end{split}  
\end{equation}  
\noindent where $\mathbb{E}$ denotes expectation over task data distribution, and $\nabla_\theta$ denotes the gradient operator. The operator $\cdot$ explicitly indicates the outer product between the gradient vector and its transpose.

The diagonal entries $F^{(t)}_i$ quantify the \textbf{importance} of parameter $\theta_i$ for task $\mathcal{T}_t$: large $F^{(t)}_i$ implies that perturbing $\theta_i$ significantly degrades the model's likelihood on $\mathcal{T}_t$.
Exact computation of $F^{(t)}$ is infeasible for large models due to its $O(d^2)$ memory complexity (where $d$ is the number of parameters). Therefore, common \textbf{Diagonal Approximation} is performed to retain only diagonal elements, by assuming parameter independence, as follows.
\begin{equation}  
  \label{eq:fim_approximation}
    \hat{F}^{(t)}_i = \frac{1}{M_t} \sum_{j=1}^{M_t} \left( \frac{\partial \log p(y_j|x_j, \theta)}{\partial \theta_i} \right)^2  
\end{equation}
As shown in prior studies such as EWC, the diagonal approximation has been validated as effective; without this simplification, the computational cost would become prohibitively high for LLMs.

%% file: chapters/4-method.tex
\section{The FGGM Framework}  
\label{sec:approach}

\subsection{Overview}
\label{subsec:overview}
\begin{figure}[t]
  \centering
  \includegraphics[scale=0.55]{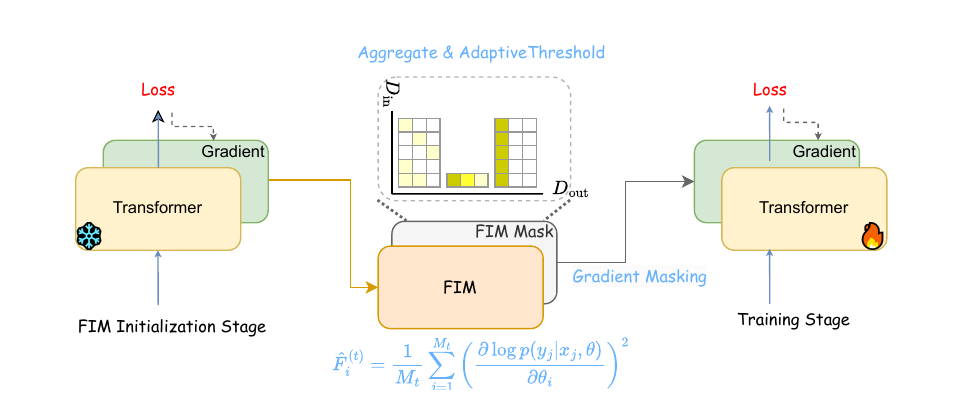}
  \caption{Illustration of the FGGM training pipeline.}
  \label{fig: pipeline}
\end{figure}

As shown in Figure~\ref{fig: pipeline}, FGGM accomplishes continual learning through a two-phase paradigm. In the first phase, the Fisher information for new datasets is estimated and \textbf{mask initialization} is conducted. In the second phase, the estimated Fisher information guides \textbf{gradient-constrained adaptation} 
so that only a subset of critical parameters is updated to prevent excessive modifications to parameters. This methodology ensures a dynamic equilibrium between stability and plasticity of model parameters-effectively preserving robustness for acquired knowledge while maintaining sufficient update flexibility for assimilating novel concepts.

\subsection{Fisher-Guided Mask Initialization}
\label{subsec:mask_initialization}

For a new task $\mathcal{T}_t$, FGGM estimates diagonal 
Fisher components 
$\hat{\mathbf{F}}^{(t)} \in \mathbb{R}^d$ 
to identify high-sensitivity parameters. Given 
training data $\{(x_j, y_j)\}_{j=1}^{M_t}$ for $\mathcal{T}_t$, empirical diagonal FIM is approximated with Eq.~\ref{eq:fim_approximation}, where gradients are computed over cross-entropy loss. 

The continuous Fisher scores undergo binarization via a threshold. Let $Q(\hat{\mathbf{F}}^{(t)}, \tau)$ 
denote the $\tau$-th quantile of empirical Fisher values. The binary mask $\mathcal{M}_i$
is defined element-wise as:  
\begin{equation}  
\mathcal{M}_j^{(t)} = 
    \begin{cases}
        1 & \text{if } \hat{F}^{(t)}_j > Q(\hat{\mathbf{F}}^{(t)}, 1 - \alpha), \\
        0 & \text{otherwise},
    \end{cases}
\end{equation}
$\alpha$ denotes masking rate (e.g., $\alpha=0.7$ retains top 30\% parameters). This adaptive thresholding adapts to non-stationary parameter importance distributions across layers, unlike fixed global thresholds.

\subsection{Mask Normalization via Output Dimension}  
\label{subsec:mask_normalization}

The mask normalization accounts for neural network parameter layout differences. For example, in fully-connected layers with weight matrix $W \in \mathbb{R}^{D_{\text{out}} \times D_{\text{in}}}$ (row-major), we aggregate FIM values across input connections per output neuron:
\begin{equation}
    \hat{F}^{(t)}_r = \sum_{c=1}^{D_{\text{in}}}\hat{F}^{(t)}_{r,c}, \quad r \in \{1,\ldots,D_{\text{out}}\},
\end{equation}
where $\hat{\mathbf{F}}^{(t)} \in \mathbb{R}^{D_{\text{out}} \times D_{\text{in}}}$ 
is the FIM estimate for weight matrix $W$.
This spatial aggregation provides two key benefits: \textbf{(1) Functional Unit Preservation}: Integrated Fisher scores represent the total sensitivity of each output neuron's computation path, protecting functionally cohesive parameter groups aligned with the network architecture. \textbf{(2) Noise Resistance}: Summation mitigates masking instability from fragmented high-dimensional FIM patterns (e.g., hidden layers), where individual weights may exhibit volatile importance signals across tasks.
For 1D parameters like biases ($\mathbf{b} \in \mathbb{R}^{D_{\text{out}}}$), this step is intentionally omitted to maintain atomic parameter resolution, as their Fisher scores directly reflect localized sensitivity without cross-connection dependencies.

The selection of \textbf{Input-dimension Aggregation (IA)} strategy is rooted in the inherent architectural properties and functional requirements of feedforward computation. Consider a fully-connected layer with weight matrix $W \in \mathbb{R}^{D_{\text{out}} \times D_{\text{in}}}$ where each output neuron's activation $y_r = \sigma(\sum_{c=1}^{D_{\text{in}}} W_{r,c}x_c + b_r)$ forms an independent computational pathway. The $r$-th row vector $W_r$ fully determines the nonlinear mapping from all input features to a single output activation. Summation over columns preserves each output neuron's functional identity as an atomic computational unit, mirroring gradient propagation patterns. Alternative \textbf{Output-dimension Aggregation (OA)} would disperse feature-specific sensitivity signals across outputs. 
Section~\ref{sec:experiments} empirically validates this analysis.

\subsection{Gradient Projection}  
\label{subsec:gradient_projection}
For task $\mathcal{T}_t$, during backpropagation, raw gradients $\mathbf{g}^{(t)}_i$ are masked as:  
\begin{equation}
\tilde{\mathbf{g}}^{(t)}_i = \mathbf{g}^{(t)}_i \odot \mathcal{M}^{(t)},
\end{equation}
where $\odot$ denotes element-wise multiplication. This projection restricts updates to parameters marked by important masks, thereby freezing non-critical regions of the parameter space. By synergizing dynamic parameter importance assessment with Fisher-guided gradient projection, FGGM achieves an equilibrium between new task acquisition and old task preservation capabilities, as empirically verified in our experiments.

%% file: chapters/5-experiments.tex
\section{Experiments}
\label{sec:experiments}
\begin{table}[t]%[!hbt]
  \caption{
    Performance comparison between FGGM and the baselines, all in 1.5B model scale, in terms of  \textit{General}  (aggregate average of MMLU, BBH, TydiQA, BoolQ, PIQA, and GSM8K task-specific metrics), \textit{TRACE Overall Performance (TRACE-OP)}.
    \textbf{IA} and \textbf{OA} denotes Input-dimension Aggregation and Output-dimension Aggregation, respectively.
  }
  \label{tab:results}
  \centering
  \setlength{\tabcolsep}{3.0pt}
  \resizebox{.95\linewidth}{!}{
  \begin{tabular}{lccc}
       \toprule
    \textbf{Method} & \textbf{Access OriData} & \textbf{General$\uparrow$} & \textbf{TRACE-OP$\uparrow$} \\
  \midrule
     ORI   & -- & 52.45 & 31.19 \\ \midrule
     EWC   & \checkmark & 50.81 & 49.69 \\ 
     REP   & \checkmark & 53.66 & \textbf{51.64} \\ \midrule
     SFT   & $\times$ & 50.89 & 49.22 \\ 
     LORA  & $\times$ & 49.48 & 43.02 \\ 
     MIGU  & $\times$ & {55.21} & 44.08 \\ 
     FGGM \textbf{(Ours)}  & $\times$ & \textbf{55.75} & {46.00} \\
     \quad w/o IA & $\times$ & 54.85	& 45.72 \\
     \quad Apply OA & $\times$ & 53.16 & 43.15 \\
   \bottomrule
  \end{tabular}
  }
\end{table}

\subsection{Experimental Setup}
\noindent \textbf{Datasets.}
 Followed MIGU~\cite{DBLP:conf/emnlp/DuCLQHCC024}, we conduct two sets of experiments. The first experiment is the \textbf{comprehensive main experiment to evaluate the overall stability and plasticity performance} of various methods, yet requiring non-trivial computational costs, while the second experiment serves an auxiliary \textbf{lightweight} experiment to thoroughly evaluate the \textbf{forgetting degree of the model as the training epoch increases}.
 For the first experiment, we evaluate learning capabilities (plasticity) of various methods on training downstream tasks on TRACE~\cite{wang2023trace}, a commonly used, comprehensive CL benchmark for LLMs. TRACE's design combines novel domain coverage avoiding instruction-tuning overlaps, 
 multidimensional evaluation spanning specialist knowledge acquisition, multilingual adaptation, code generation accuracy, and justified numerical reasoning, with standardized metrics ensuring full reproducibility across tasks.
 Following \cite{wang2023trace},  evaluating retention of the General capabilities (stability) is conducted across six principal dimensions using established benchmarks: MMLU, BBH, TyDiQA, PIQA, BoolQ, and GSM8K.
 For the second experiment, to analyze the impact of training dynamics on model performance, following the MIGU work~\cite{DBLP:conf/emnlp/DuCLQHCC024}, we incorporate Magicoder-Evol-Instruct-110K~\cite{DBLP:conf/icml/0003W0D024}, for code generation tasks. To assess code acquisition capabilities, we employ the Humaneval benchmark~\cite{DBLP:journals/corr/abs-2107-03374} for evaluating \textit{plasticity} and the BBH benchmark to evaluate \textit{stability}.
 
\noindent \textbf{Baselines.}
  We compare against six established CL approaches, implemented on Qwen2~\cite{DBLP:journals/corr/abs-2407-10671}:
  \textbf{(1) ORI}: The original pre-trained model without fine-tuning.
  \textbf{(2) EWC}~\cite{DBLP:journals/corr/KirkpatrickPRVD16}: Implements elastic weight consolidation via Fisher information regularization on critical parameters.  
  \textbf{(3) REP}: Experience Replay that preserves historical samples and periodically replays them with new data.
  \textbf{(4) SFT}: Sequential Fine-Tuning with unconstrained parameter updates using cross-entropy loss.  
  \textbf{(5) LORA}~\cite{DBLP:conf/iclr/HuSWALWWC22}: Trains low-rank decomposition matrices while freezing base parameters for lightweight adaptation.  
  \textbf{(6) MIGU}~\cite{DBLP:conf/emnlp/DuCLQHCC024}: Selectively freezes trainable weights based on activation magnitude thresholds to mitigate forgetting\footnote{The MIGU paper~\cite{DBLP:conf/emnlp/DuCLQHCC024} compares MIGU with the soft-masking DAS method on the DAS benchmark and demonstrates that the DAS method exhibits weaker plasticity than MIGU.}.

\noindent \textbf{Implementation Details.}
  $\alpha$ is set to the same $0.7$ as used in MIGU for fair comparison.
  Following TRACE~\cite{wang2023trace}, the learning rate is initialized to $1 \times 10^{-5}$ and decayed linearly to $0$, except for LORA, where the learning rate is set to $1 \times 10^{-4}$. Following MIGU, we adopt the default sequential training order: C-STANCE, FOMC, MeetingBank, Py150, ScienceQA, NumGLUE-cm, NumGLUE-ds, and 20Minuten, with epochs set to 5, 3, 7, 5, 3, 5, 5, and 7, respectively. AdamW~\cite{DBLP:conf/iclr/LoshchilovH19} is employed for optimization. All experiments are run on NVIDIA Tesla A100 80G GPUs. Brain floating-point format ($\text{BF16}$)~\cite{DBLP:journals/corr/abs-1905-12322} is applied to accelerate the training and decoding processes. All code is implemented in PyTorch.

\noindent \textbf{Evaluation Metrics.}
  The evaluation metrics are task-specific and include Accuracy, F1 Score, ROUGE-L~\cite{lin-2004-rouge}, BLEU~\cite{DBLP:conf/acl/PapineniRWZ02}, and SARI~\cite{DBLP:journals/tacl/XuNPCC16}. 
  Accuracy measures the proportion of correctly predicted samples in a classification task.
  F1 Score, the harmonic mean of precision and recall, provides a balanced evaluation.
  ROUGE-L assesses the similarity between generated text and reference text by evaluating the longest common subsequence.
  BLEU calculates the n-gram overlap between the generated text and reference translations.
  SARI is designed for text simplification tasks, evaluating the quality of additions, deletions, and preservations in the system output relative to reference simplifications and the original input. We use two primary metrics,  \textbf{General}  (aggregate average of MMLU, BBH, TydiQA, BoolQ, PIQA, and GSM8K task-specific metrics), and \textbf{TRACE Overall Performance (TRACE-OP)}.
  Following training on the $t^{th}$ task, where $RD_{t,i}$ denotes post-task performance on the $i^{th}$ task ($i \geq t$), the Overall Performance (OP)~\cite{DBLP:conf/eccv/ChaudhryDAT18} of TRACE averages $\text{OP}_t$ across all learned tasks, where
    $\text{OP}_t = \frac{1}{t} \sum_{i=1}^{t} RD_{t,i}$
  Evaluations are all conducted using OpenCompass~\cite{2023opencompass}. 

\subsection{Results and Analysis}

\textbf{Main Results.}  Table~\ref{tab:results} presents a comprehensive comparison of our proposed FGGM against the baselines across multiple evaluation benchmarks, all in 1.5B model size. As can be seen, \textbf{\textit{FGGM achieves the best general performance (stability) among all methods}}.
  For methods that allow access to old task data, since they can utilize historical TRACE task data during training, it is understandable that they achieve substantially better performance on TRACE-OP.
  Since no method in Table~\ref{tab:results} can access the data from the General evaluation sets, the performance of REP and EWC on General is markedly inferior to FGGM. 
  It is important to point out that in real-world applications, historical data is often unavailable due to privacy concerns; hence, the advantages of REP and EWC on downstream tasks will diminish.

\begin{table}[t]%[!hbt]
  \caption{
    FGGM performance w.r.t. $\alpha$ in 1.5B model size.
  }
  \label{tab: alpha_results}
  \centering
  \setlength{\tabcolsep}{3.0pt}
  \resizebox{.7\linewidth}{!}{
  \begin{tabular}{ccc}
   \toprule
    \textbf{Value of $\alpha$} & \textbf{General$\uparrow$} & \textbf{TRACE-OP$\uparrow$} \\
   \midrule
     0.5  & 53.81 & 49.56 \\
     0.6  & 53.92 & 48.23 \\
     0.7  & 55.75 & 46.00 \\
     0.8  & 56.21 & 44.85 \\
     0.9  & 57.65 & 40.12 \\
   \bottomrule
  \end{tabular}
  }
\end{table}

\noindent \textbf{Ablation of Dimension Aggregation.}
To assess the impact of our mask normalization strategy, we compare the full model with two variants: one without input dimension aggregation (w/o IA) and another applying output dimension aggregation instead (Apply OA).
As shown in the last three rows of Table~\ref{tab:results}, the full FGGM model outperforms the two variants on both General and TRACE-OP performance.
Removing input dimension aggregation (w/o IA) degrades General from 55.75 to 54.85, suggesting that \textbf{\textit{input dimension aggregation plays a supportive role in the normalization process}}.
In contrast, applying output dimension aggregation (Apply OA) leads to a more significant decline, with General dropping to 53.16 and TRACE-OP declining to 43.15. 
Although output dimension aggregation theoretically offers similar noise resistance to input dimension aggregation, it disrupts the consistency of the model’s output parameters, leading to performance degradation.

To further understand the effects of dimension aggregation, we compute similarity scores between tasks in the TRACE dataset.
The similarity scores using output aggregation exhibit higher inter-task similarities, especially between semantically related tasks.
However, this aggregation strategy also leads to reduced differentiation between tasks, 
indicating that output aggregation may blur the distinctions between tasks. This homogenization can result in the model struggling to retain distinct knowledge for each task, thereby increasing forgetting. Conversely, with input aggregation,
while there are still notable similarities between related tasks,
the overall range of similarity scores is more balanced, suggesting that input dimension aggregation preserves the unique aspects of each task while still capturing relevant inter-task relationships. This balance allows the model to preserve task-specific sensitivities more effectively, facilitating better knowledge retention and reducing interference between tasks.
 Overall, the ablation study demonstrates that \textbf{\textit{input dimension aggregation is essential for preserving the functional integrity of output neurons and ensuring robust continual learning performance, whereas output dimension aggregation adversely affects both performance}}.

\noindent \textbf{Impact of the masking rate $\alpha$.}
Hyperparameter $\alpha$ is to control the proportion of parameters retained for stability. We set $\alpha=0.7$ in all experiments for a fair comparison with MIGU. 
From Table~\ref{tab: alpha_results}, we observe that, as expected,  more parameter updates (smaller $\alpha$) lead to lower stability (lower General), while simultaneously bringing greater plasticity and resulting in higher TRACE-OP.

  \begin{figure}[t]
    \centering
    \subfigure[Performance on HumanEval]{%
        \includegraphics[scale=0.16]{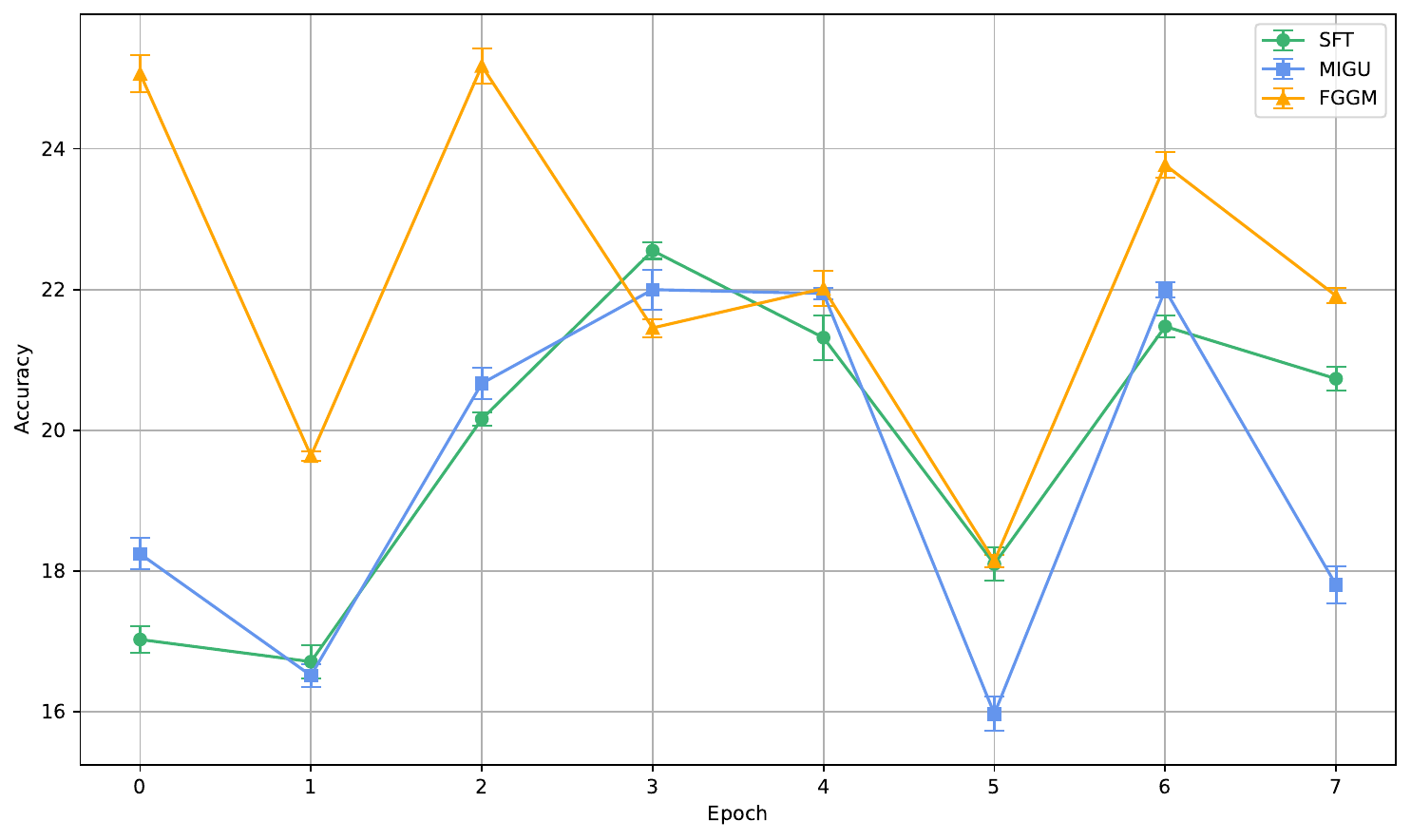}
        \label{fig:comparison_acc}
        }
    \subfigure[Performance on BBH]{%
        \includegraphics[scale=0.16]{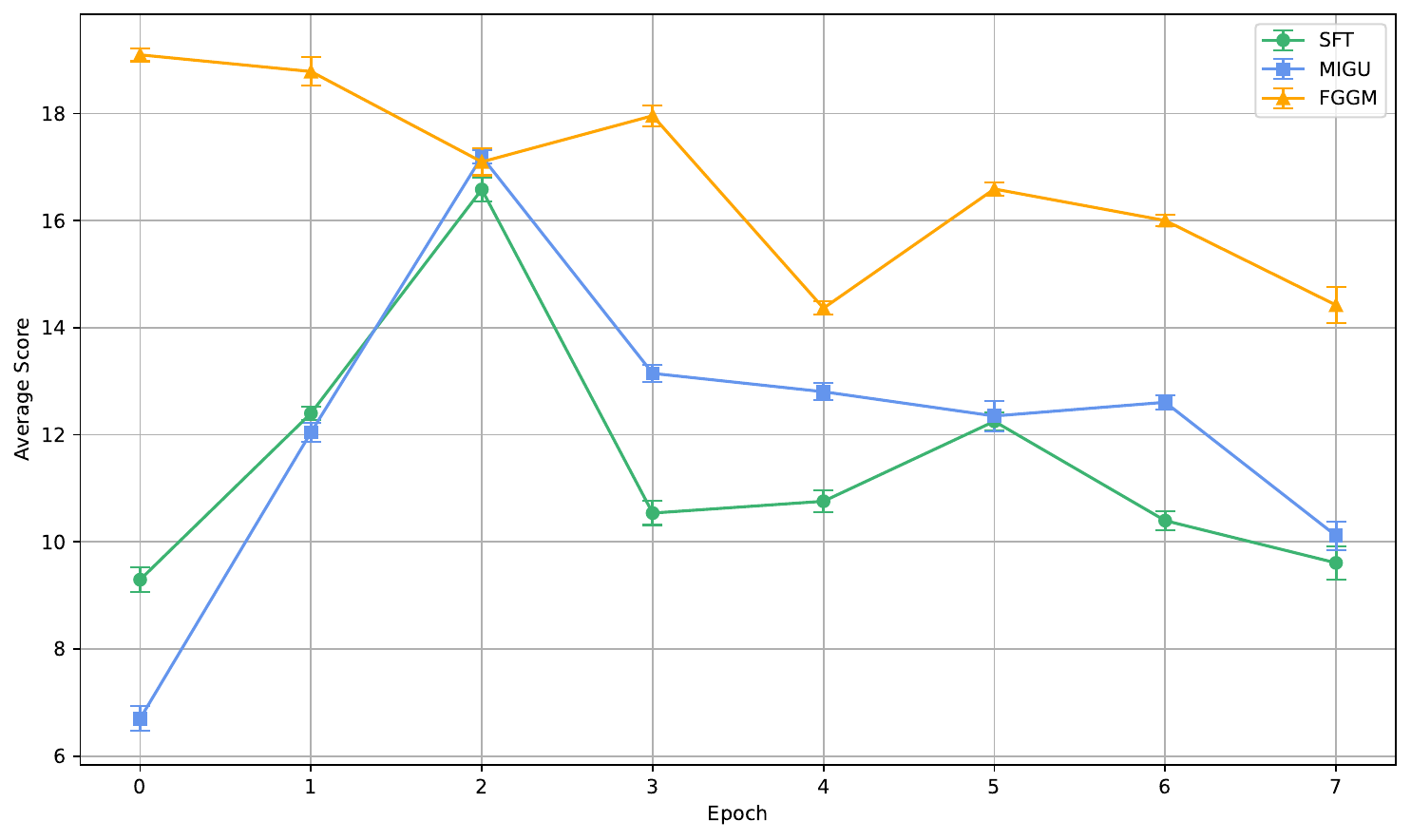}
        \label{fig:comparison_plot}
    }
    \caption{Performance comparison between SFT, MIGU, and our FGGM on the HumanEval and BBH benchmarks through progression of training epoches, from training on Magicoder-Eval-Instruct-110K, all in 0.5B model size.}
    \label{fig:fggm_training_pipeline}
  \end{figure}
  
  \noindent \textbf{Learning Same, Forgetting Less.}
  Figure~\ref{fig:fggm_training_pipeline} shows the performance of training on Magicoder-Eval-Instruct-110K and evaluating on the HumanEval and BBH benchmarks. 
  Due to the high computational cost of the experiment, we conduct only on the 0.5B model size.
  Figure~\ref{fig:comparison_acc} shows the \textbf{HumanEval} results, assessing the \textbf{model's learning ability}. FGGM consistently achieves higher accuracy across all eight evaluation metrics compared to SFT and MIGU. Specifically, FGGM scores range from 17.07 to 25.00, demonstrating a significant improvement in learning capability. Figure~\ref{fig:comparison_plot} shows the performance on the \textbf{BBH} benchmark, assessing the \textbf{model's forgetting process}. Here, FGGM maintains higher performance 
  compared to SFT and MIGU: SFT and MIGU exhibit lower scores, indicating greater forgetting. 
  FGGM consistently outperforms both SFT and MIGU, suggesting that FGGM not only enhances learning efficiency but also effectively mitigates forgetting. Overall, the results demonstrate that Magicoder-Eval-Instruct-110K with FGGM training achieves superior learning outcomes on HumanEval while simultaneously exhibiting reduced forgetting on BBH, suggesting that \textbf{\textit{FGGM provides a balanced improvement in both acquisition and retention of knowledge, making it a robust training strategy for maintaining performance over time}}.

  \begin{table}[t]%[!hbt]
  \caption{Impact of model size scaling.
  }
  \label{tab: scaling_up}
  \centering
  \setlength{\tabcolsep}{3.0pt}
  \resizebox{.75\linewidth}{!}{
  \begin{tabular}{lccc}
   \toprule
    \textbf{Model} & \textbf{Size} & \textbf{General$\uparrow$} & \textbf{TRACE-OP$\uparrow$} \\
   \midrule
     MIGU & 1.5B & 55.21 & 44.08 \\
     FGGM & 1.5B & {55.75} & {45.99} \\ \midrule
     MIGU & 7B   & 68.92 & 58.66 \\
     FGGM & 7B   & {70.83} & {60.77} \\
   \bottomrule
  \end{tabular}
  }
  \end{table}

 \noindent \textbf{Scaling up Model Size.} Table~\ref{tab: scaling_up} shows that FGGM consistently outperforms MIGU on both General and TRACE-OP metric, based on both Qwen2-1.5B and Qwen2-7B architectures. Particularly on the General metric, the gain from FGGM over MIGU widens progressively as model size increases.
 These results demonstrate FGGM's better scalability compared to baseline methods.

\noindent \textbf{Computational costs.}
\label{appendix:computational_costs}
Methods such as OGD~\cite{DBLP:conf/aistats/FarajtabarAML20} that involve gradient orthogonality require retaining the vector space of the entire dataset and projecting the current model’s vectors onto this space, resulting in prohibitively high computational costs and making these methods infeasible for LLM applications with memory constraints. FGGM does not need to maintain all vectors, thus the memory consumption is affordable. FGGM introduces two additional overheads compared to SFT: FIM computation and Gradient Masking.
(1) We employ offline FIM computation and accelerate the process using batch processing. To further enhance efficiency of FIM computation, we can transition from offline FIM to online FIM.
(2) Gradient Masking only involves masking the backpropagated gradients, hence this operation is computationally efficient.

\noindent \textbf{Soft-Masking versus Hard-Masking.}
Soft-masking applies continuous suppression weights (e.g., 0–1 coefficients) to gradients, offering balanced plasticity-stability tradeoffs. 
It preserves parameter adaptability through partial updates, but risks gradual forgetting under conflicting task gradients and requires careful importance estimation. 
In contrast, hard-masking ensures
strict stability, yet may overly restrict model plasticity when tasks share substructures. 
To preserve stability and minimize computational overhead, FGGM adopts the hard-masking approach.